\documentclass[letterpaper, 10 pt, conference]{ieeeconf}  

\IEEEoverridecommandlockouts                              

\usepackage{cite}
\usepackage{mdwmath}
\usepackage{mdwtab}
\usepackage[caption=false]{subfig}
\usepackage{url}

\usepackage{graphics} 
\usepackage{epsfig} 
\usepackage{mathptmx} 
\DeclareMathAlphabet{\mathcal}{OMS}{cmsy}{m}{n}
\usepackage{times} 
\usepackage{amsmath} 
\usepackage{amssymb}  

\usepackage[utf8]{inputenc}
\usepackage{cleveref}
\usepackage{booktabs}
\usepackage{adjustbox}
\usepackage{threeparttable}
\usepackage{pifont}

\usepackage[nolist,nohyperlinks]{acronym}
\newacro{CNN}{Convolutional Neural Network}
\newacro{DNN}{Deep Neural Network}
\newacro{GPS}{Global Positioning System}
\newacro{GNSS}{Global Navigation Satellite System}
\newacro{NLOS}{non-line-of-sight}
\newacro{ADAS}{Advanced Driver Assistance Systems}
\newacro{LIDAR}[LiDAR]{Light Detection And Ranging}
\newacro{HD map}{High Definition map}
\newacro{EV}{Embedding Vector}
\newacro{SLAM}{Simultaneouos Localization And Mapping}
\newacro{MLP}{MultiLayer Perceptron}
\newacro{IMU}{Inertial Measurement Unit}
\newacro{ML}{Machine Learning}
\newacro{SfM}{Structure from Motion}
\newacro{PnP}{Perspective-n-Points}
\newacro{ASPP}{Atrous Spatial Pyramid Pooling}

\def\etal{\emph{et al. }}
\def\eg{\emph{e.g., }}
\def\ie{\emph{i.e., }}
\def\wrt{\emph{w.r.t. }}

\title{\LARGE \bf
Global visual localization in LiDAR-maps through shared 2D-3D embedding space
}

\author{D. Cattaneo$^{1,2}$, M. Vaghi$^{1}$, S. Fontana$^{1}$, A. L. Ballardini$^{1,3}$, D. G. Sorrenti$^{1}$
\thanks{$^{1}$ Universit\`a degli Studi di Milano - Bicocca, Milano, Italy}%
\thanks{$^{2}$ University of Freiburg, Freiburg, Germany}%
\thanks{$^{3}$ Universidad de Alcal\'a, Alcal\'a de Henares, Spain}%
\thanks{Contributions according to~\cite{ira-authorship-roles}: conceptualization DC, development DC, data curation SF, drafting DC+MV, revising all authors, supervision ALB+DGS}%
\thanks{The work of ALB while at Universidad de Alcal\'a has been funded by EU Horizon 2020 grant Marie Sk\l{}odowska-Curie n. 754382}%
}

\usepackage{tikz}
\usepackage{textcomp}
\usepackage{lipsum}

\newcommand\copyrighttext{%
  \footnotesize \textcopyright 2019 IEEE. Personal use of this material is permitted. Permission from IEEE must be obtained for all other uses, in any current or future media, including reprinting/republishing this material for advertising or promotional purposes, creating new collective works, for resale or redistribution to servers or lists, or reuse of any copyrighted component of this work in other works.}
\newcommand\copyrightnotice{%
\begin{tikzpicture}[remember picture,overlay]
\node[anchor=south,yshift=10pt,xshift=10pt] at (current page.south) {\fbox{\parbox{\dimexpr\textwidth-\fboxsep-\fboxrule\relax}{\copyrighttext}}};
\end{tikzpicture}%
}

\begin{document}

\maketitle
\copyrightnotice 

\thispagestyle{empty}
\pagestyle{empty}

\begin{abstract}
Global localization is an important and widely studied problem for many robotic applications. Place recognition approaches can be exploited to solve this task, \eg in the autonomous driving field. While most vision-based approaches match an image \wrt an image database, global visual localization within LiDAR-maps remains fairly unexplored, even though the path toward high definition 3D maps, produced mainly from LiDARs, is clear. In this work we leverage \ac{DNN} approaches to create a shared embedding space between images and LiDAR-maps, allowing for image to 3D-LiDAR place recognition. We trained a 2D and a 3D \ac{DNN} that create embeddings, respectively from images and from point clouds, that are close to each other whether they refer to the same place. An extensive experimental activity is presented to assess the effectiveness of the approach \wrt different learning paradigms, network architectures, and loss functions. All the evaluations have been performed using the Oxford Robotcar Dataset, which encompasses a wide range of weather and light conditions.
\end{abstract}

\section{INTRODUCTION}\label{sec:introduction}
Estimating the pose of a robot \wrt a map, \ie \emph{localization}, is a fundamental part of any mobile robotic system. It has such relevance also for autonomous cars, which usually operate in very large and dynamic environments, where this task is particularly challenging.

A typical localization pipeline usually 
involves two steps.
First, a rough pose of the robot is estimated, thus performing a \emph{global} localization of the observer. Secondly, the initial rough pose is refined, and an accurate pose is estimated, therefore performing a \emph{local} localization. Usually, the first step can be accomplished with the aid of \acp{GNSS}, which provides a global position. However, the localization 
reliability of \acp{GNSS} is inadequate
, particularly in urban environments, where buildings might block or deflect the signals from the satellites, leading to \ac{NLOS} and multi-path issues.

Aiming at overcoming \acp{GNSS} limitations, many different methods to solve the localization problem have been proposed, including vision-based and \ac{LIDAR}-based approaches.
Although \ac{LIDAR} sensors are the de facto standard for commercial large 3D geometric reconstructions, their price, weight, and mechanically-based scanning systems, which are not rated viable for the automotive sector, make them still not suitable for mass installation on cars. 
\begin{figure}[t]
  \begin{center}
  \includegraphics[width=.99\columnwidth]{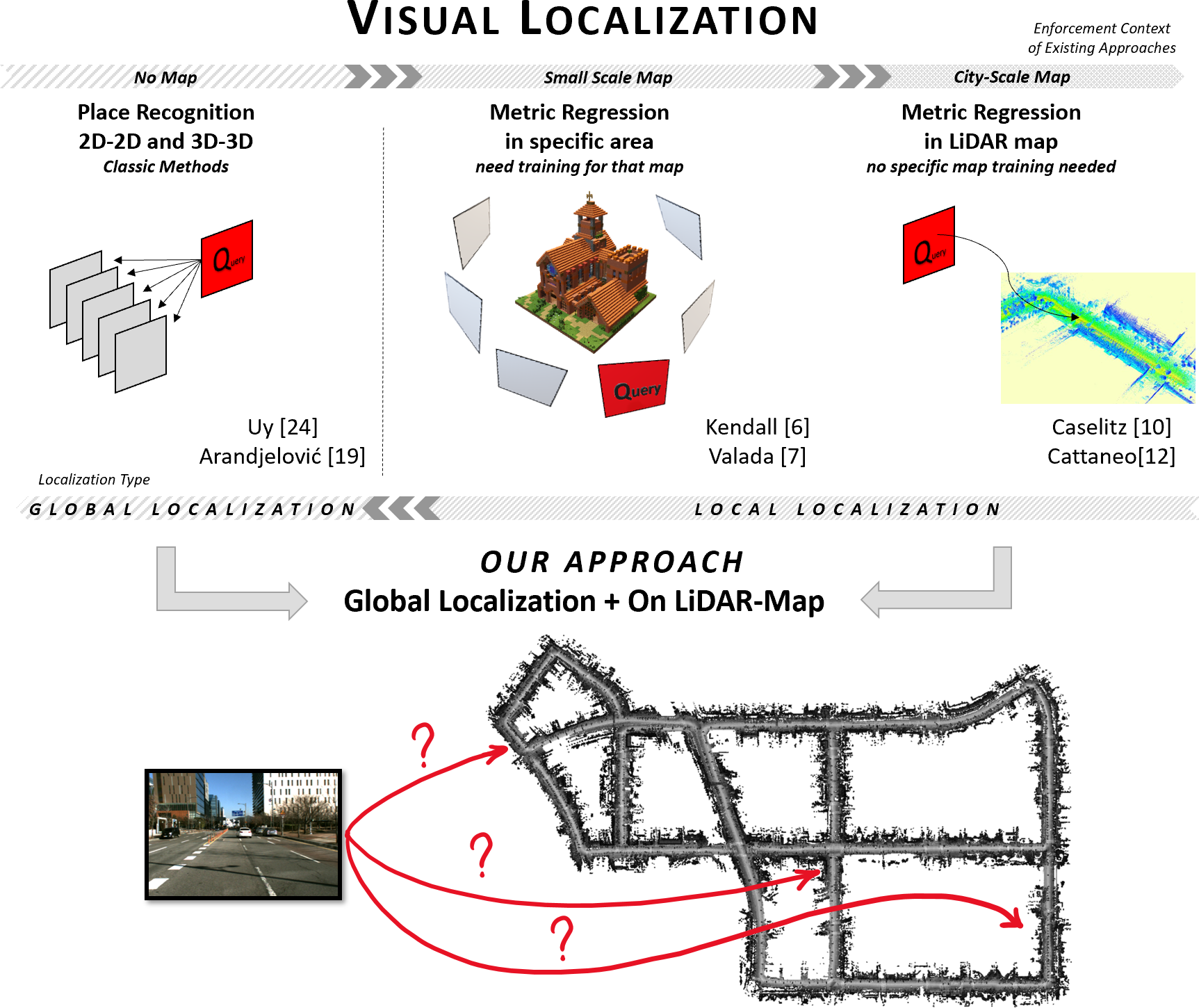}
  \end{center}
  \caption{A short overview of the related existing localization methods.
  Our proposal enables for the first time a global localization from an image \wrt a city-scale 3D map, typically obtained from LiDARs.
  This allows the usage only of cheap automotive-grade cameras on-board the vehicle, and the soon-available commercial HD grid maps.}
  \label{fig:figure_1_teaser}
\end{figure}
Some approaches in the literature~\cite{Ballardini_ITSC_2016, Ballardini_ICRA_2019, Tao2013, Schreiber2013} tackle the local localization task by matching high level features extracted from images from on-board camera(s), \eg buildings, road intersections, and/or road markings, with data available from online mapping services, \eg OpenStreetMap.

Recent \ac{ML} approaches, mainly based on \acp{CNN} or random forests, solve the localization problem in a single step by regressing the robot's pose using only a single RGB image~\cite{Kendall_2015_ICCV,8458420, Shotton_2013_CVPR}. However
, they require a huge amount of images of the environment to train the model, and this problem is particularly relevant in outdoor environments (\eg wide urban areas).
Moreover, these approaches 
learn to perform localization only \wrt the specific places used during the training phase. Therefore, for new locations, the acquisition of a new dataset, together with a re-training stage is required.
This characteristic makes these methods not suitable for autonomous road driving.

Nowadays, map-making companies, \eg HERE and TomTom, are investing large amounts of resources in developing so-called HD maps. These maps are designed specifically for the automotive domain, and provide accurately localized high-level features, such as traffic lights, and road signs. An evolution of these maps includes also accurate 3D large geometric reconstructions, in the form of point clouds. Therefore, it is quite likely that, in the future, these 3D HD maps will be used by autonomous vehicles, for more effective and safer systems~\cite{seif2016autonomous}.

While approaches that exploit the same sensor for both localization and mapping usually achieve good performances, localizing a vehicle \wrt a \ac{LIDAR}-map exploiting only visual information is a difficult task, due to the different nature of the sensors used for mapping and for localization.
The difficulty of this cross-media task is confirmed by the relative absence of specific literature.
The few approaches to tackle the visual localization task \wrt \ac{LIDAR}-maps fall in one of two categories. On the one hand, the 3D geometry of the scene, as reconstructed from cameras, can be matched to the 3D map~\cite{Caselitz_2016}. On the other hand, the match can be performed in the image space, by comparing camera images to the expected projection of the map onto a virtual image plane, by means of Normalized Mutual Information (NMI)~\cite{Wolcott_2014} or \ac{CNN}-based techniques~\cite{Cattaneo_2019}.
Both categories tackle the local localization task, \ie they need a rough estimate of the pose.

\begin{figure}[t]
  \begin{center}
  \includegraphics[width=.99\columnwidth]{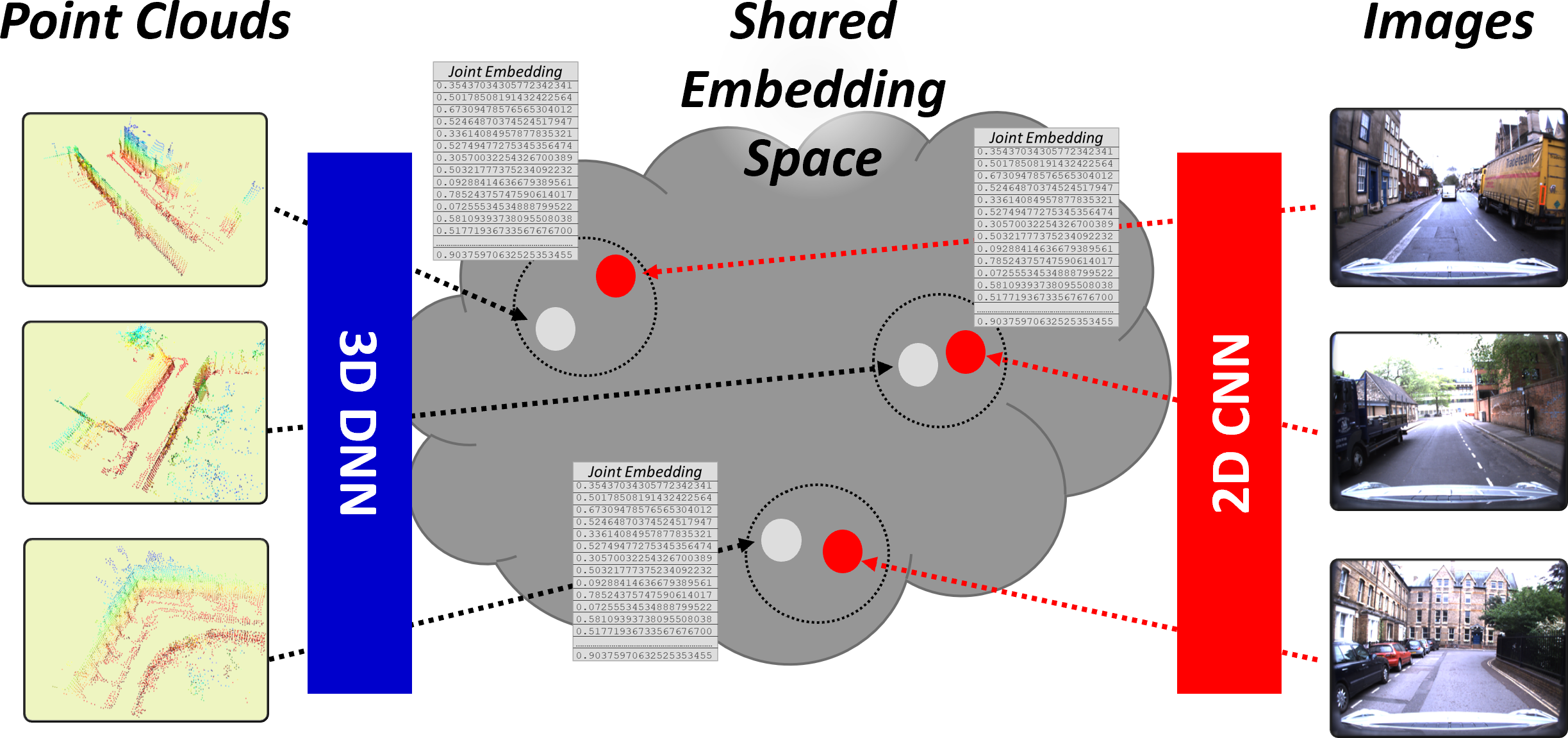}
  \end{center}
  \caption{We leverage DNN-based approaches to create a shared embedding space between images and LiDAR point clouds, allowing for image to 3D-LiDAR place recognition.}
  \label{fig:embeddingspace}
\end{figure}

How to perform global visual localization \wrt \ac{LIDAR}-map,s when \acp{GNSS} are not available, is still an open question, see~\Cref{fig:figure_1_teaser}.
In this paper we propose a novel \ac{DNN}-based approach to the global visual localization problem, \ie to localize the observer of an RGB image in a previously built geometric 3D map of an urban area.
Specifically, we propose to jointly train two \acp{DNN}, one for the images and the other for the point clouds, in order to generate a shared embedding space for the two types of data, see~\Cref{fig:embeddingspace}.
This embedding space allows us to perform place recognition with heterogeneous sensors: an image query \wrt a point cloud database is just one of the possibilities.
Besides allowing localization on a geometric map using visual information, our proposal does not need to be trained for every new environment.
Therefore, our approach can associate an image, \ie the query, to the corresponding point cloud, among those that constitute the map of the area, \ie the database.

The paper is organized as follows: Section \ref{sec:related-work} describes the existing approaches for place recognition. In Section \ref{sec:proposed-approach} we present the details of our approach. In Section \ref{sec:experimental-results} we present and discuss the obtained experimental results. Finally in Sections \ref{sec:conclusion}, we shortly draw some conclusions.

\section{RELATED WORKS}\label{sec:related-work}
Historically, visual place recognition represents a relevant problem for the computer vision community. The aim of this task is to decide whether and where a place, observed in an image, has already been observed within a set of geo-referenced images. This is relevant in mobile robotics, as it may serve as a starting point for local localization systems, or to detect loop closures in \ac{SLAM} systems~\cite{williams2009comparison, newman2005slam}.

Traditional methods to perform place recognition were based on approaches such as bag of words, which exploit handcrafted-features (\eg SIFT and SURF~\cite{Lowe2004, SURF2006}), visual vocabulary~\cite{Phiblin_2007_CVPR}, and query expansion~\cite{Chum_2011_CVPR}. More recently, methods that exploit deep learning techniques have been proposed. As an example, Arandjelovic \etal \cite{Arandjelovic_2016_CVPR} exploited a CNN-based feature extractor and proposed a specific pooling layer, named NetVlad, designed for place recognition, which was inspired by the Vector of Locally Aggregated Descriptors~\cite{jegou2010}.
The definition of discriminative descriptors represents a fundamental step of the place recognition task.
Within \ac{ML} methodologies, the definition of descriptors task is named \textit{metric learning}; the resulting feature space is called embedding space, and the descriptors take the name of \acfp{EV}.
A possible \ac{ML} method, named \textit{contrastive}, provides sample pairs to a model by considering two cases: either the two samples represent the same concept or distinct ones~\cite{Hadsell_CVPR2006}.
This training method tries to minimize the distance between \acp{EV} of the same class while enforcing the distances between negative samples to be higher than a threshold.
Another similar technique consists in comparing triplets of samples, which are organized such that a given input is associated with a positive sample and a negative one. 
In particular, positive samples represent the same concept of the input, while the negative samples belong to a different one. Different approaches extended the triplet embedding method by considering multiple negative samples at once instead of a single one~\cite{Song_2016_CVPR,NIPS2016_6200}.

Although place recognition is a task traditionally reserved to vision-based systems, recently Uy \etal \cite{Uy_2018_CVPR} proposed a \ac{DNN}-based approach that performs place recognition on \ac{LIDAR}-maps.
This method, named PointNetVLAD, matches an input point cloud, \eg acquired from a vehicle, to another one among those contained in a database representing a large urban area. It can match (the point cloud of) a \ac{LIDAR} scan to a location in a \ac{LIDAR}-map of a large urban area. PointNetVlad uses PointNet~\cite{Qi_2017_CVPR} as feature extractor, and NetVlad~\cite{Arandjelovic_2016_CVPR} to compute the descriptors.

To the best of our knowledge, the only work in the literature that treats the problem of image and \ac{LIDAR}-map descriptors matching, using a metric learning approach, has been proposed by Feng \etal \cite{Mengdan2019}. In their work a neural network, named \textit{2D3D-MatchNet}, performs descriptors extraction from 2D image and 3D \ac{LIDAR} patches. The objective is to perform the metric localization of a camera through 2D to 3D patch matching. Although a neural network is used to compute the descriptors, 2D key-points are detected from images using the detector in the SIFT software, while 3D key-points are extracted from the point clouds using Intrinsic Shape Signatures~\cite{Zhong_2009}. The key-points are then used to define the patches used for computing the descriptors, from images and \ac{LIDAR}-maps. The localization is finally computed solving a \ac{PnP} problem. Their work is limited also in that they do not handle \ac{LIDAR}-maps whose extension is larger than a few tens of meters, which is not global localization.

There are many other contributions in the literature, which match image descriptors to 3D maps, but they refer to maps produced by \ac{SfM} approaches~\cite{ozyecsil2017survey}, 
where an image descriptor is attached to every point in the map.
On the other hand, we need to handle descriptor-less point clouds, as naturally coming from \ac{LIDAR} sensors.

\section{PROPOSED APPROACH}\label{sec:proposed-approach}
Differently from visual place recognition techniques, our approach matches a single RGB image to a database of known-pose point clouds, to retrieve the point cloud representing the same place of the query image. In order to compare images and point clouds, we propose to learn a shared embedding space where data representing the same place live close to each other even though they have to be computed from different types of data. In particular, we propose two \acp{DNN}, one for the images and another for the point clouds, jointly trained to produce similar \acp{EV}, when the image and the point cloud come from the same place.

Our work was inspired by those of Uy \etal \cite{Uy_2018_CVPR} and Feng \etal \cite{Mengdan2019}. Differently from~\cite{Uy_2018_CVPR}, which requires a \ac{LIDAR} on-board the vehicle, our approach only requires a camera on-board, which is very common in modern cars. Differently from~\cite{Mengdan2019} our approach is designed to work on city-scale maps.
More formally, given a query image $\mathcal{I}$ and a \ac{LIDAR}-map $\mathcal{M}$, we can split the map into multiple overlapping sub-maps $m_i$. We formulate the global localization task as a metric learning problem: we want to find two mapping functions $f(\cdot)$ and $g(\cdot)$, implemented as two \acp{DNN} such that $d(f(\mathcal{I}), g(m_i)) < d(f(\mathcal{I}), g(m_j))$ when $m_i$ represent the same place where $\mathcal{I}$ was taken, and $m_j$ does not. $d(\cdot)$ is a distance function, such as the euclidean distance. The domain of $f(\cdot)$ is the image space ($\mathbb{R}^{HxWx3}$, Height $x$ Width $x$ 3 channels: R, G, and B), while the domain of $g(\cdot)$ is the point cloud space ($\mathbb{R}^{Nx3}$, $N$ points $x$ 3 coordinates: $X$, $Y$, $Z$). The output of both functions is an \ac{EV} of fixed length $K$. 
Once we have found the mappings between the input spaces and the \ac{EV} space, to compute a global localization we just need to compare the \ac{EV} of the query image with the \acp{EV} of all the sub-maps $m_i \in \mathcal{M}$.

\subsection{2D Network Architecture}
The network that computes the \acp{EV} from images is composed of two parts. First, a \ac{CNN} extracts local features from the input image; then those features are aggregated to provide a fixed length \ac{EV}. Concerning the \ac{CNN}, we considered some of the most relevant architectures for the image classification task. In particular, we tested the following networks: VGG-16~\cite{simonyan2014very} and ResNet-18~\cite{He_2016_CVPR}. 
Since we are not interested in the classification task, we removed the fully connected layers from the latter architectures. 
For the aggregation step, we tested two different methods: the NetVlad layer~\cite{Arandjelovic_2016_CVPR}, that was specifically designed for the place recognition task, and a simple \ac{MLP}.

\subsection{3D Network Architecture}
Similarly to the 2D \ac{CNN} architecture, for the 3D network we adopted the same approach: a \ac{DNN}-based features extractor followed by a features aggregation layer. The best data structures and models that allow the extraction of discriminative descriptors from 3D point clouds are not clear in the literature, therefore a comparison between different architectures was needed. The first 3D feature detector considered was PointNet~\cite{Qi_2017_CVPR}, since it was the first 3D \ac{DNN} approach to directly process point clouds as a list of points. We also considered one interesting extension of PointNet, named EdgeConv~\cite{Wang_2018}.
The latter two \acp{DNN} require point clouds composed of a fixed number of points, thus a down-sampling step of the input is necessary. The last option we considered is the feature extraction layer of SECOND~\cite{YAN_2018}. Since we are not interested in classification or segmentation tasks, we modified the networks as follows:
\begin{itemize}
    \item PointNet: we cropped the network before the MaxPooling layer;
    \item EdgeConv: we cropped the segmentation network after the concatenation layer;
    \item SECOND: we replaced the RPN with an \ac{ASPP} layer~\cite{chen2017deeplab}
\end{itemize}
Finally, for the aggregation step, we used again the NetVlad layer and a \ac{MLP}.

These models accommodate different representations of the input data (\eg unordered lists~\cite{Qi_2017_CVPR}, graphs~\cite{Wang_2018}, voxel grids~\cite{YAN_2018}), thus we think that providing an evaluation of the existing approaches is an important step forward for this type of research.

We propose two different learning paradigms to create a shared embedding space between the 2D-3D \acp{EV}: \textit{teacher/student}, and \textit{combined}.

\subsection{Teacher/Student Training}
Given a 2D and a 3D neural network, the teacher/student method first trains one of the models (the teacher) to create an effective embedding space for a particular task. In a second step, this pre-trained network will be used to help the second one (the student) to also generate a similar embedding space, \ie similar \acp{EV} for similar concepts. For instance, initially a \ac{CNN} model learns to perform place recognition with images, then a 3D \ac{DNN} tries to emulate the \ac{CNN} descriptors. In this way, the network creates the embedding space where \acp{EV} are defined and the student tries to also generate that space from a different kind of data.

In this work, we first train the 2D network with a triplet loss function~\cite{facenet}: given a triplet $(I^a_i, I^p_i, I^n_i)$ composed of an anchor image $I^a_i$, an image depicting the same place $I^p_i$ (\emph{positive}) and an image of a different place $I^n_i$ (\emph{negative}), the loss function is defined as:
\begin{equation}
    \mathcal{L}_{trp}^{2D{\text{-to-}}2D} = \sum_i [d(f(I^a_i), f(I^p_i)) - d(f(I^a_i), f(I^n_i)) + m]_+
    \label{eq:triplet}
\end{equation}
where $d(\cdot)$ is a distance function, $m$ is the desired separation margin, $f(\cdot)$ is the 2D network we want to train, and $[\cdot]_+$ means $Max(0, [\cdot])$.
Once the 2D network (teacher) has been trained, the 3D network (student) was trained to mimic the output of the teacher (to obtain a Joint Embedding). In this case, given a pair $(I_i, m_i)$ composed of an image $I_i$ and a point cloud $m_i$ captured at the same time, the loss function is:
\begin{equation}
    \mathcal{L}^{JE} = \sum_{i} d(f(I_i), g(m_i))
    \label{eq:je}
\end{equation}
Please note that, during this step, only the student network $g(\cdot)$ is trained, while the teacher $f(\cdot)$ is kept fixed. 


\subsection{Combined Training}
Alternatively, we propose a combined approach that simultaneously trains both the 2D and the 3D neural networks, in order to produce the same embedding space. In this case, the loss function proposed to jointly train both networks is composed of different components.

\noindent
\textit{Same-Modality metric learning.}
The first components of the proposed combined loss are aimed at producing effective \acp{EV} for place recognition within the same modality, \ie the same type of sensor data (query image \wrt image database and query point cloud \wrt point cloud database). The same-modality loss is defined as follow:
\begin{equation}
    \mathcal{L}_{trp}^{SM} = \mathcal{L}_{trp}^{2D{\text{-to-}}2D} + \mathcal{L}_{trp}^{3D{\text{-to-}}3D}
\end{equation}
Here, $\mathcal{L}_{trp}^{2D{\text{-to-}}2D}$ is the 2D-to-2D triplet loss defined in \cref{eq:triplet}, while the 3D-to-3D loss $\mathcal{L}_{trp}^{3D{\text{-to-}}3D}$ can be derived similarly.

\begin{figure}
  \begin{center}
  \includegraphics[width=.99\columnwidth]{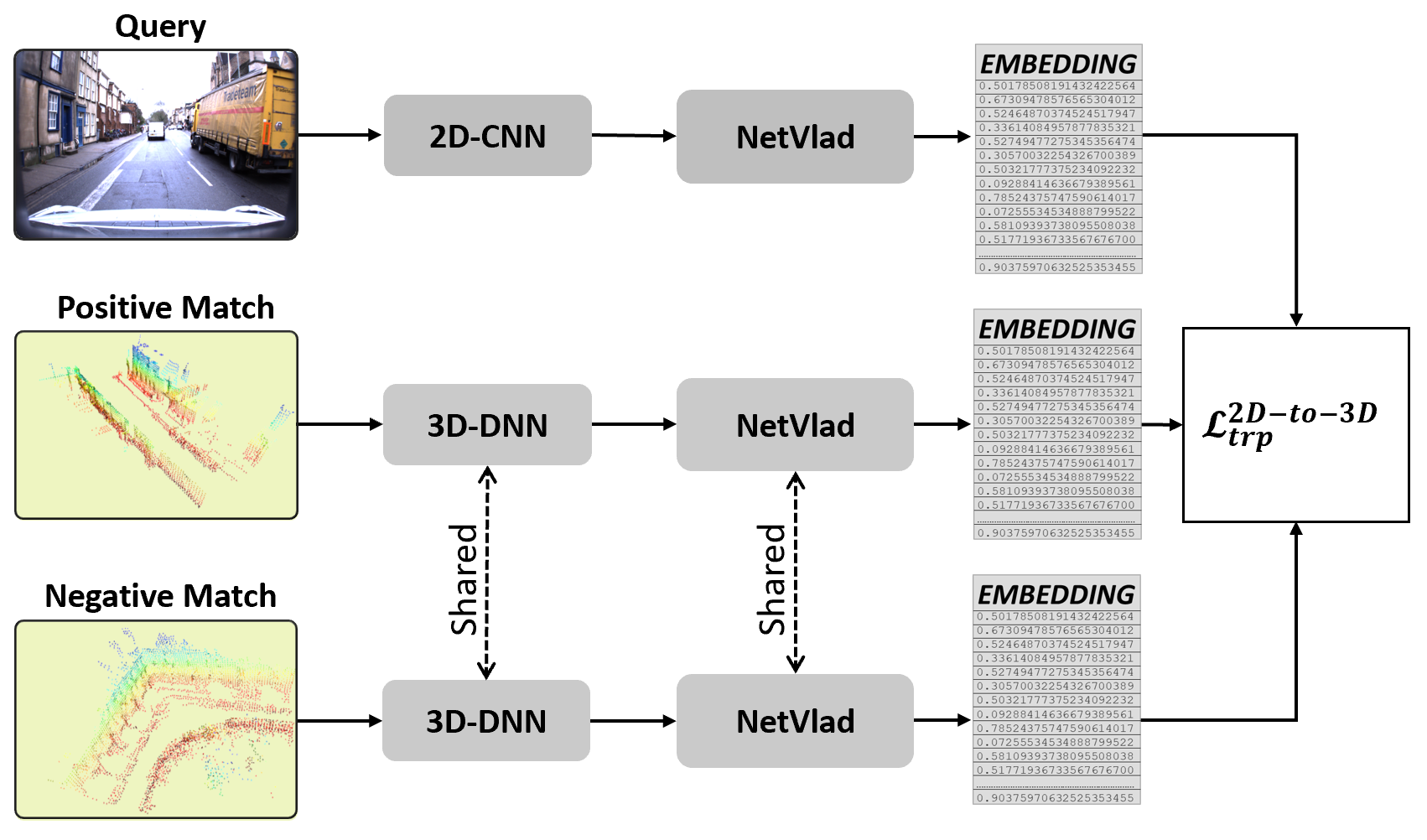}
  \end{center}
  \caption{
    During the training phase, the ``triplet'' technique considers a positive and a negative sample with respect to a query.
    Please note that the weights of the 3D-DNN and the associated NetVLAD layers are shared between the point cloud samples, meaning that we use the same network.}
  \label{fig:triplet}
\end{figure}

\noindent
\textit{Cross-Modality metric learning.}
In order to learn 2D and 3D \acp{EV} that live in the same embedding space, we extend the triplet loss to perform cross-modality metric learning:
\begin{align}
    \mathcal{L}_{trp}^{2D{\text{-to-}}3D} &= \sum_i [d(f(I^a_i), g(m^p_i)) - d(f(I^a_i), g(m^n_i)) + m]_+ \\
    \mathcal{L}_{trp}^{3D{\text{-to-}}2D} &= \sum_i [d(g(m^a_i), f(I^p_i)) - d(g(m^a_i), f(I^n_i)) + m]_+ \\
    \mathcal{L}_{trp}^{CM} &= \mathcal{L}_{trp}^{2D{\text{-to-}}3D} + \mathcal{L}_{trp}^{3D{\text{-to-}}2D}
\end{align}
An example of the loss computed on a triplet is depicted in \Cref{fig:triplet}.

\noindent
\textit{Joint Embedding loss.}
The last component of the proposed loss tries to minimize the distance between 2D and 3D \acp{EV} recorded at the same time, \ie we want the \ac{EV} of a point cloud to be as close as possible to the \ac{EV} of the corresponding image. To achieve this aim we used the joint embedding loss defined in \cref{eq:je}; however, in this case both networks $f(\cdot)$ and $g(\cdot)$ are trained.




\noindent
\textit{Full combined loss.}
The final loss used to jointly train the 2D and 3D networks is a combination of the aforementioned components ($\lambda_1=0.1, \lambda_2=1, \lambda_3=1$):
\begin{equation}
    \mathcal{L}_{total} = \lambda_1 \mathcal{L}_{trp}^{SM} + \lambda_2 \mathcal{L}_{trp}^{CM} + \lambda_3 \mathcal{L}^{JE}
\end{equation}

\subsection{Training details}
During the training phase, to mitigate over-fitting, we use the following data augmentation scheme: for the images, we first apply a random color jitter (changing the saturation, brightness, hue, and contrast). 
Then, we randomly rotate the image within a range of [$-5^\circ$, $+5^\circ$]; finally, we apply a random translation on both axes, with a maximum value of 10\% of the size of the image on that axis. 
Regarding the point clouds, we applied a random rigid body transformation, with a maximum translation of 1.5 meters on all the axes, and a maximum rotation of $10^\circ$ around the vertical axis and $2^\circ$ around the lateral and longitudinal axes. We chose these values to accommodate slightly different points of view.

Moreover, we applied random horizontal mirroring to the images and, when an image is mirrored, the relative point cloud is also mirrored. This particular augmentation is applied before any other data augmentation. 

For the implementation we used the PyTorch library~\cite{paszke2017automatic}.
The input images are undistorted and resized to 320x240.
The batch size is constructed by randomly selecting $N=4$ places and randomly picking two samples of each of the selected places (for a total of $8$ images and $8$ point clouds per batch), so to have $N$ positive pairs within the batch. The negative samples are selected randomly (one for each positive pair) within the batch.

\subsection{Inference}
Once the \ac{DNN} models have been trained, the inference of the place recognition is pursued as follows. First, the 3D \ac{LIDAR}-map of the environment is split into overlapping sub-maps and the 3D \ac{DNN} is used to generate the corresponding \acp{EV}. These \acp{EV} are then organized in a specific data structure in order to allow a subsequent fast nearest neighbor querying, we used a KD-tree approach. Lastly, the localization is performed by comparing the \ac{EV} of the query image, generated by the 2D \ac{CNN}, \wrt the database. It would of course be possible to reverse the paradigm, \ie to produce a dataset of 2D image \acp{EV} and then perform queries using the \acp{EV} from 3D sub-maps or \ac{LIDAR} scans.

\section{EXPERIMENTAL RESULTS}\label{sec:experimental-results}

This section describes the experiments performed to evaluate the different architectures and approaches, including the dataset preprocessing task.

\subsection{Dataset}
In order to develop an approach that is robust to challenging environmental conditions (\eg scene structure, lights, and seasonal changes), we worked on the well known Oxford RobotCar~\cite{RobotCarDatasetIJRR} dataset. RobotCar contains geo-referenced LiDAR scans, images, and \ac{GPS} data of urban areas. These data were obtained by traversing the same path multiple times over a year.
It includes a large set of weather conditions, which allowed us to develop a robust approach. For instance, it is possible to test the place recognition performance by considering different light conditions for images, and structural changes for the scene, which are reflected in corresponding changes in the \ac{LIDAR}-maps, \eg in the presence of roadworks. RobotCar has also been used for training and testing 2D3D-MatchNet~\cite{Mengdan2019} and PointNetVlad~\cite{Uy_2018_CVPR}, which inspired our research.

\subsection{Regions Subdivision and Sub-maps Creation}
We considered a total of 43 runs, the same used in \cite{Uy_2018_CVPR}. For each run, an image is stored every five meters and the corresponding \ac{LIDAR} sub-map is cropped from the whole map.
For each 3D sub-map we considered a range of 50 meters on each axis, and we removed the ground plane.
The training and validation sets have been created by performing a subdivision of the global path in different non-overlapping regions.
The goal here was to make it possible to evaluate the performance in regions never provided to the neural networks during the training phase. In particular, we used the same four regions defined in~\cite{Uy_2018_CVPR}.

\subsection{Evaluation metrics}
The evaluation of a place recognition approach is usually based on the \emph{recall@k} measure.
The term $k$ represents the number of places that our system determines to be the most similar \wrt the input query. If at least one of the retrieved $k$ elements corresponds to the location of the input query, then the retrieval is considered correct.
When the number of samples in the database increases, the difficulty of the operation increases as well, therefore we fixed $k$ to the $1\%$ of the samples contained in the database in order to ensure the invariance of the measure with respect to the database size. In our experiments, we considered two poses to belong to the same place if they are spaced less than $20m$, as in~\cite{Uy_2018_CVPR}.

To test the performance of our approach we proceeded as follows. For every possible pair of distinct runs, again in~\cite{Uy_2018_CVPR}, we used all samples of the first run as database, and only the samples within the validation area of the second run as queries. In this way, we tested our method in places never seen during the training phase. Finally, we compute the mean of the recall@k metric over all pairs.
Note that the recall@1\% metric is computed for every run-pairs, based on the number of samples of the database (ranging from 96 to 402), and then averaged on all pairs.

\subsection{Results}

We leveraged the networks proposed in~\cite{Arandjelovic_2016_CVPR} (VGG16+NetVLAD), and a modified version of~\cite{YAN_2018} (SECOND+ASPP+NetVLAD) in order to extract \acp{EV} from both 2D and 3D data.
We applied the teacher/student learning paradigm with the smooth-L1 distance function~\cite{girshick2015fast}, fixing the 2D architecture as teacher, since it represents the state of the art in the place recognition task with images.
In particular, we used the pre-trained version of the 2D backbone (in particular, VGG16 trained on TokyoTM) and performed fine-tuning on the Oxford Robotcar dataset. 

\begin{table*}[ht]
\setlength\tabcolsep{5pt}
\centering
\begin{threeparttable}
\caption{Best model vs all runs}
\label{tab:best-model}
\begin{tabular}{cccc|cccc|cccc}
\toprule
\multicolumn{4}{c}{\textbf{recall@1\%}}                                  & \multicolumn{4}{c}{\textbf{recall@1}}                                    & \multicolumn{4}{c}{\textbf{recall@5}}                                    \\ \cmidrule(lr){1-4} \cmidrule(lr){5-8} \cmidrule(lr){9-12}
2D-to-2D    & 3D-to-2D    & 2D-to-3D & 3D-to-3D     &2D-to-2D    & 3D-to-2D    & 2D-to-3D       & 3D-to-3D  &2D-to-2D    & 3D-to-2D    & 2D-to-3D       & 3D-to-3D  \\ \midrule
96,63 & 70,44 & 77,28            & 98,43            & 88,40             & 29,51           & 41,92            & 93,99            & 95,76             & 54,79           & 64,34            & 97,45  \\ \bottomrule         
\end{tabular}
    \begin{tablenotes}[para,flushleft]
      \footnotesize      
      \item In this table we show the retrieval performances of our approach, computed over all the pairs of runs. We report the recall@1\%, recall@1, and recall@5, in all the four possible modalities, \eg 2D-to-3D represents 2D queries \wrt 3D database.
    \end{tablenotes}
\end{threeparttable}
\vspace{-2mm}
\end{table*}

The results of these experiments are provided in \Cref{tab:best-model}, where we show the performance \wrt all the possible query-database combinations. Even though our work is mainly focused on the 2D-to-3D modality, we also obtained comparable results in the 2D-to-2D place recognition modality, and even state-of-the-art performances for 3D-to-3D. Considering the novelty of the proposed approach, the obtained results are promising.
The lower recall achieved in the 3D-to-2D modality might be explained considering that the two learned embedding spaces (from images and point clouds respectively) capture not only the similar traits from the two media, but also distinct and media-specific traits, which are not easily discernible.

An important aspect concerns the recall achieved when performing 3D queries on 3D database: we found that our approach, which exploits SECOND, achieved higher recall@1 than other approaches~\cite{Uy_2018_CVPR,pcan,lpd_net} on the same task, as shown in \Cref{tab:comapare-pvd}.
To perform a comparison we followed an evaluation scheme similar to theirs, considering images and point clouds with an interval of $20m$ (instead of $5m$) for the database runs and $10m$ for the query runs, over the same 43 runs. However, it is fair to note that those networks take in input a fixed number of points (4096), while our approach does not make any assumption on the number of points.
Moreover, their sub-maps have dimension $25m$ on each axis, while ours have dimension $50m$.

\begin{table}[ht]
\fontsize{6.3}{7.2}\selectfont
\setlength\tabcolsep{1.9pt}
\centering
\begin{threeparttable}
\caption{3D Place Recognition Comparison}
\label{tab:comapare-pvd}
\begin{tabular*}{\linewidth}{lcccc|cccc}
\toprule
      & \multicolumn{4}{c}{\textbf{recall@1\%}}&  \multicolumn{4}{c}{\textbf{recall@1}} \\
    \cmidrule(lr){2-5}  \cmidrule(lr){6-9}
      & 2D-to-2D  & 3D-to-2D  & 2D-to-3D  & 3D-to-3D  & 2D-to-2D  & 3D-to-2D  & 2D-to-3D  & 3D-to-3D  \\ \midrule
PNV    & -      & -      & -      & 80.09           & -       & -      & -      & 63.33          \\
PCAN   & -      & -      & -      & 86.40           & -       & -      & -      & 70.72          \\
LPD    & -      & -      & -      & \textbf{94.92}  & -       & -      & -      & 86.28          \\ 
Ours   & 89.79  & 49.52  & 63.95  & 93.24           & 79.57   & 26.81  & 39.65  & \textbf{87.56} \\ \bottomrule
\end{tabular*}
    \begin{tablenotes}[para,flushleft]
      \footnotesize      
      \item Retrieval performances comparison between our approach, PointNetVLAD~\cite{Uy_2018_CVPR}, PCAN~\cite{pcan} and LPD-Net~\cite{lpd_net}.
      Both recall@1\% and recall@1 are reported.
    \end{tablenotes}
\end{threeparttable}
\vspace{-2mm}
\end{table}

Finally, we performed an ablation study to investigate the effect of different components of the system.
In particular, we tested the system without the data augmentation techniques (mirroring and point clouds transformation), and without the \ac{ASPP}.
In these tests, we only considered a subset of 10 runs, to have a quick insight on the effect of the components. The results, shown in \Cref{tab:res-voxlenet}, show that both data augmentation, and the \ac{ASPP} bring substantial improvements to the performance.

\begin{table}[ht]
\scriptsize
\setlength\tabcolsep{2.6pt}
\centering
\begin{threeparttable}
\caption{Ablation Study}
\label{tab:res-voxlenet}
\begin{tabular*}{\linewidth}{lccc|cccc}
\toprule
& & & & \multicolumn{4}{c}{\textbf{Modality}} \\ \cmidrule(lr){5-8}
\textbf{Test}    & \textbf{ASPP}         & \textbf{Mirroring}    & \textbf{PC Augm.}         & 2D-to-2D & 3D-to-2D             & 2D-to-3D             & 3D-to-3D             \\ \midrule
Base                & \ding{55} & \ding{55} & \ding{55} & 96.67                  & 61.53          & 73.17          & 96.92          \\
1                   & \ding{51} &\ding{55} & \ding{55} & 96.67                  & 65.31          & 73.99          & 97.75          \\
2                   & \ding{51} & \ding{51} & \ding{55} & 96.67                  & 66.30          & 76.76          & 97.66          \\
3                   & \ding{51} & \ding{51} & \ding{51} & 96.67                  & \textbf{71.10} & \textbf{79.10} & \textbf{98.44} \\
\bottomrule
\end{tabular*}
    \begin{tablenotes}[para,flushleft]
      \footnotesize      
      \item Retrieval performances of our approach in terms of recall@1\% by varying different components of the system.
    \end{tablenotes}
\end{threeparttable}
\vspace{-2mm}
\end{table}

\begin{figure}[bh!]
\centering
    \subfloat{%
    \includegraphics[width=.49\columnwidth]{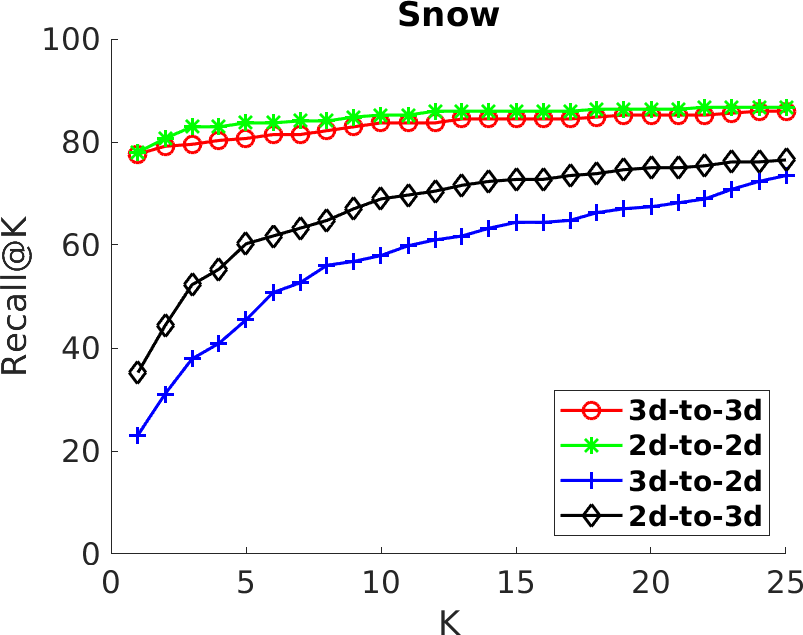}}
    \subfloat{%
    \includegraphics[width=.49\columnwidth]{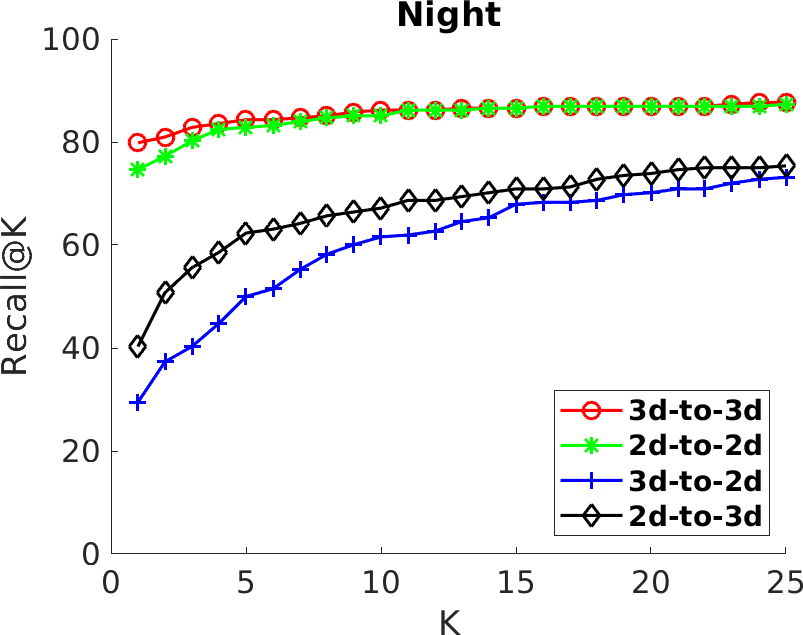}}\\
    \subfloat[]{%
    \includegraphics[width=.49\columnwidth]{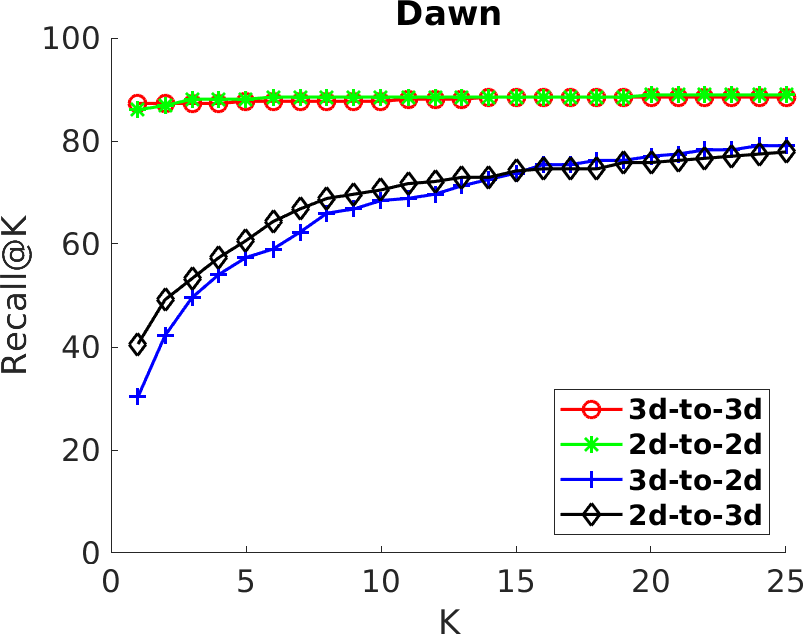}}
    \subfloat[]{%
    \includegraphics[width=.49\columnwidth]{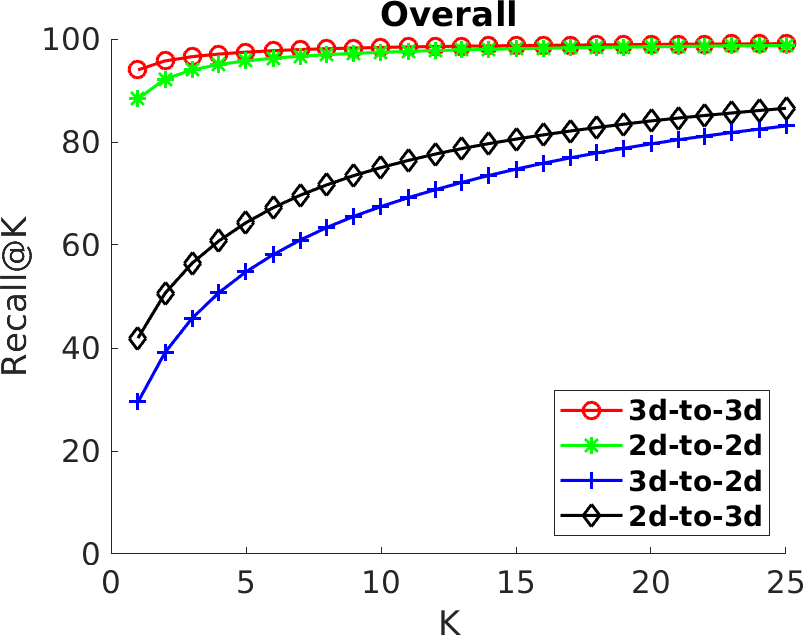}}
    \caption{Recall measures considering up to $K=25$ similar places. Here 
    challenging weather and time conditions, \ie Snow, Night and Dawn, were compared against an ``overcast'' database. We also provide the plot of the average recall using all available dataset runs.}
    \label{fig:matlab}
\end{figure}

Furthermore, once we found the combination which provides the best results, we challenged the robustness of our approach by considering different weather conditions between \acp{EV} database and the queries provided to the trained system. For instance, we generated a database from 3D samples gathered during on overcast run and then we provide to the model images acquired during winter. 
This approach has also been exploited by considering different light conditions.
In \Cref{fig:matlab} we show the recall@k of the four modalities with different values of $k$.
In particular, a query run with snow conditions (run \textit{2015-02-03-08-45-10}), one recorded during nighttime (run \textit{2014-12-16-18-44-24}) and one recorded during dawn (run \textit{2014-12-16-09-14-09}) is compared against a ``overcast'' database (run \textit{2015-10-30-13-52-14}).
The average recall using all available dataset runs is also reported.


\subsection{Further Improvement Attempts}
Having reached satisfactory results, we tried to increase our model's performance by varying different sub-components of the system.
Starting from the architecture type, we tested different backbones such as VGG-16 and ResNet18 for what concerns the 2D data, and PointNet, EdgeConv and SECOND for the 3D data. 
Furthermore, we investigated various learning methods together with different loss functions and distance measures.
Despite our best efforts, the results in \Cref{tab:res-old} show that these variations did not improve the performance of the model, instead they reduced the recall capability of our system.


\begin{table}[ht]
\scriptsize
\setlength\tabcolsep{3pt}
\centering
\begin{threeparttable}
\caption{Recall@1\% varying different sub-components}
\label{tab:res-old}
\begin{tabular*}{\linewidth}{cccc|cccc}
\toprule
\multicolumn{2}{c}{\textbf{Backbone}} &&& \multicolumn{4}{c}{\textbf{Modality}} \\ \cmidrule(lr){1-2} \cmidrule(lr){5-8}
2D    & 3D   & Loss      & Dist.   & 2D-to-2D & 3D-to-2D & 2D-to-3D & 3D-to-3D \\ \midrule
Resnet18 & PointNet & (1)        & MSE    & 51.13    & 20.61    & 9.93     & 95.30    \\
Vgg16    & PointNet & (1)        & MSE    & 85.51    & 31.98    & 51.32    & 94.53    \\
Vgg16    & PointNet & (2)     & MSE    & 86.62    & 38.77    & 53.46    & 95.61    \\
Vgg16    & PointNet & (2)    & Cosine & 88.03    & 36.85    & 52.16    & 96.31    \\
Vgg16    & PointNet & (2)     & L2     & 83.38    & 36.48    & 47.18    & 94.64    \\
Vgg16    & PointNet & (3) & L2     & 96.64    & 31.33    & 28.69    & 92.00    \\
Vgg16    & EdgeConv & (3) & L2     & 96.60    & 67.52    & 59.50    & 97.21    \\
Vgg16    & SECOND   & (2)     & MSE    & 89.27    & 53.90    & 59.75    & 97.10    \\ \bottomrule
\end{tabular*}
    \begin{tablenotes}[para,flushleft]
      \footnotesize      
      \item Loss function are defined as follows: (1) $\mathcal{L}_{trp}^{SM} + \mathcal{L}_{trp}^{CM}$, (2) $\mathcal{L}_{trp}^{SM} + \mathcal{L}_{trp}^{CM} + \mathcal{L}^{JE}$ and (3) is the teacher/student method.
    \end{tablenotes}
\end{threeparttable}
\vspace{5mm}
\end{table}

\section{CONCLUSIONS}\label{sec:conclusion}
In this paper, we described a novel \ac{DNN}-based approach to perform global visual localization in \ac{LIDAR}-maps. 
In particular, we proposed to jointly train a 2D \ac{CNN} and a 3D \ac{DNN} to produce a shared embedding space. 
We trained and validated our \ac{DNN} on the challenging Oxford Robotcar dataset, including all weather conditions, and promising results show the effectiveness of our approach. 
We also obtained performance comparable to state-of-the-art approaches on both 2D-to-2D and 3D-to-3D modalities, although this was not the focus of our research. 
Even though the obtained results are promising 
the cross-modality retrieval is still far from the same-modality performances, leaving room for future improvements.
To our knowledge, an approach that performs a global visual localization using \ac{LIDAR}-maps has never been presented before.

\section*{ACKNOWLEDGMENTS}
The authors would like to thank Pietro Colombo for the help in the editing of the associated video.





\bibliographystyle{IEEEtran}
\bibliography{IEEEabrv,bibliography}

\end{document}